\documentclass[letterpaper, 10 pt, conference]{ieeeconf}  

\IEEEoverridecommandlockouts                              

\usepackage{graphics} 
\usepackage{epsfig} 
\usepackage{times} 
\usepackage{amsmath} 
\usepackage{amssymb}  
\usepackage{booktabs}
\usepackage{multirow}
\usepackage{url}
\usepackage{hyperref}
\usepackage{ragged2e}

\usepackage[absolute,overlay]{textpos}

\let\labelindent\relax
\usepackage[shortlabels]{enumitem}
\title{\LARGE \bf
Zero-Shot Scene Understanding for Automatic Target Recognition Using  Large Vision-Language Models
}

\author{Yasiru Ranasinghe, Vibashan VS, James Uplinger, Celso De Melo, and Vishal M. Patel
\thanks{Yasiru Ranasinghe, Vibashan VS, and Vishal M. Patel are with the Department of Electrical and Computer Engineering, The Johns Hopkins University, Baltimore, MD, USA. Emails: 
{\tt\small \{dranasi1, vvishnu2, vpatel36\}@jhu.edu}.}%
\thanks{James Uplinger and Celso De Melo are with the DEVCOM Army Research Laboratory, Adelphi. Email: 
{\tt\small \{james.r.uplinger7.civ, celso.m.demelo.civ\}@army.mil}.}%
}

\begin{document}

\maketitle
\thispagestyle{empty}

\pagestyle{plain}

\begin{abstract}
Automatic target recognition (ATR) plays a critical role in tasks such as navigation and surveillance, where safety and accuracy are paramount. In extreme use cases, such as military applications, these factors are often challenged due to the presence of unknown terrains, environmental conditions, and novel object categories. Current object detectors, including open-world detectors, lack the ability to confidently recognize novel objects or operate in unknown environments, as they have not been exposed to these new conditions. However, Large Vision-Language Models (LVLMs) exhibit emergent properties that enable them to recognize objects in varying conditions in a zero-shot manner. Despite this, LVLMs struggle to localize objects effectively within a scene. To address these limitations, we propose a novel pipeline that combines the detection capabilities of open-world detectors with the recognition confidence of LVLMs, creating a robust system for zero-shot ATR of novel classes and unknown domains. In this study, we compare the performance of various LVLMs for recognizing military vehicles, which are often underrepresented in training datasets. Additionally, we examine the impact of factors such as distance range, modality, and prompting methods on the recognition performance, providing insights into the development of more reliable ATR systems for novel conditions and classes.
\end{abstract}

\begin{textblock*}{8cm}(11.72cm,26.58cm) 
\noindent\parbox{8cm}{\raggedleft \textit{Approved for public release: distribution is unlimited.}}
\end{textblock*}

\section{INTRODUCTION}

Automatic Target Recognition (ATR) \cite{bhanu1986automatic,novak1997automatic, patel2010automatic} is essential for modern surveillance and defense, enabling the automated detection and classification of targets in sensor data using image processing and machine learning. ATR systems provide rapid, accurate object identification in complex environments, crucial for military applications \cite{hinton2015distilling,zhang2016accelerating} where precision is vital. Beyond defense, ATR is used in autonomous driving and navigation \cite{uccar2017object, ratches2011review}, making it key for both national security and commercial automation \cite{taddeo2022artificial}.

\begin{figure}[!ht]
	\centering
	\includegraphics[width=.9\linewidth]{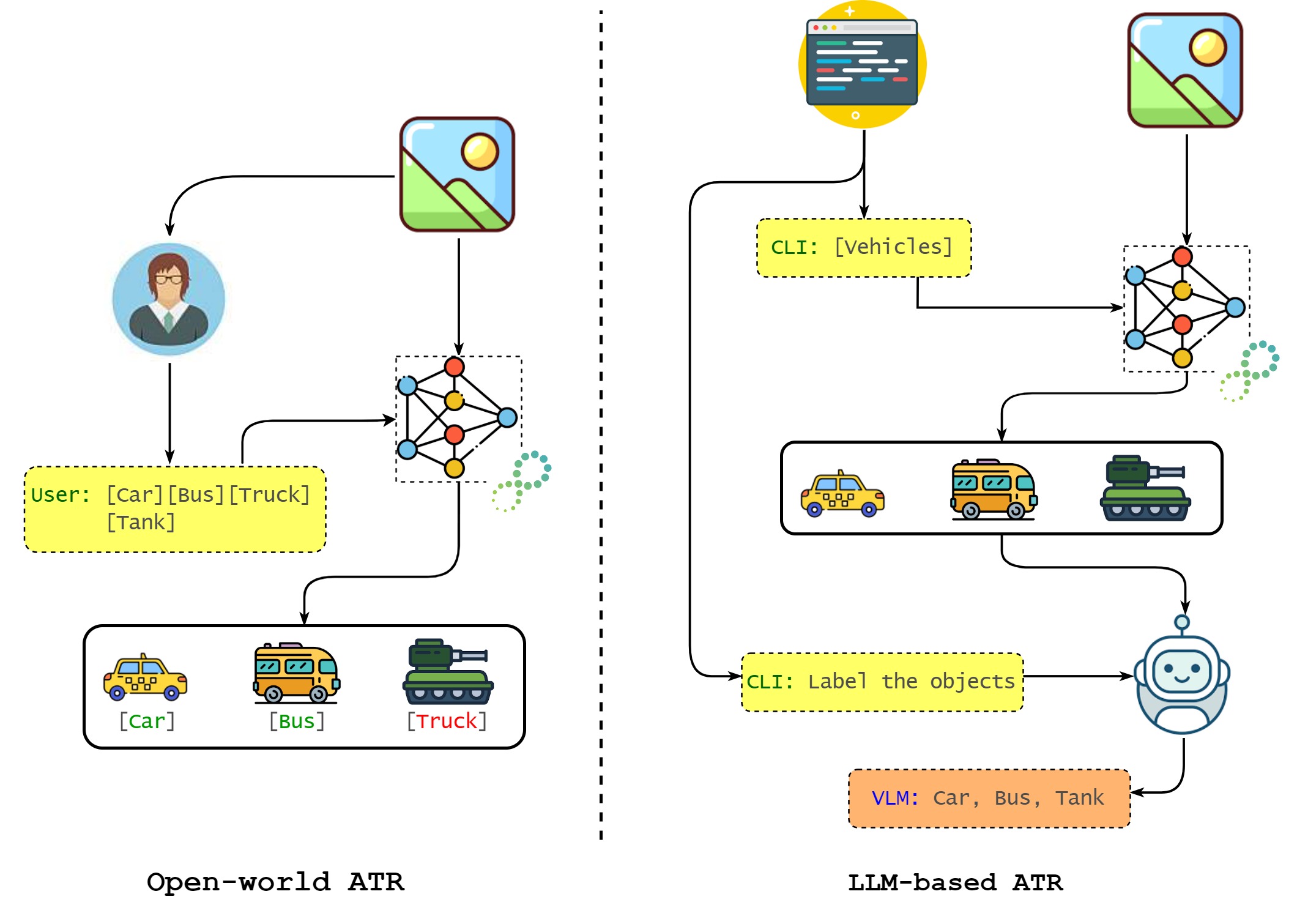}
    \vskip-13pt
    \caption{Comparison between existing architectures zero-shot text prompted automatic target recognition (ATR). Standard open-world ATR involves a human-in-the-loop as the novel objects to be detected and recognized should be provided to the detector. Even then, the state-of-the-art open-world ATR systems fail to recognize novel object classes that extremely deviate from training classes. In LLM-based ATR, the detector is only used at the capacity of localizing the objects present in the image. Then, each localized object is sent to a larger vision-language model to recognize the object, which eliminates the need for user interference.
	}
	\label{fig: headline}
\end{figure}

A reliable system for ATR is critical for ensuring the robustness and safety \cite{bhanu1986automatic,novak1997automatic} of systems deployed in dynamic and uncertain environments. Autonomous systems, such as drones or autonomous vehicles \cite{bathla2022autonomous}, rely heavily on machine learning models to identify and classify objects. However, these models are typically trained on specific datasets and may not perform well when encountering data that significantly deviates from the training distribution \cite{amodei2016concrete, hendrycks2021unsolved}. OOD detection techniques \cite{yang2024generalized, miyai2024generalized, vs2022meta} aim to identify these anomalies by measuring the uncertainty or confidence of the model's predictions. Methods such as Bayesian neural networks, which provide a probabilistic measure of uncertainty \cite{mitros2019validity}, and distance-based metrics in feature space \cite{gangireddy2023knowing}, are commonly employed to flag data points that the model finds ambiguous or unfamiliar. By detecting OOD samples, autonomous systems can be programmed to take precautionary measures \cite{brundage2018malicious}, such as requesting human intervention or switching to a more conservative decision-making mode \cite{kim2015impact}, thereby enhancing overall safety and effectiveness.

Open-world object detectors \cite{joseph2021towards, wang2021unidentified} represent a significant advancement in ATR systems by addressing the limitations of traditional models that typically operate under a closed-world assumption \cite{fei2016breaking}, where the system only recognizes previously seen classes. These open-world detectors are designed to not only identify known objects with high accuracy but also detect and categorize unknown objects as 'unknowns'. This capability is essential in dynamic environments where new object types \cite{gidaris2018dynamic} can appear without prior label data. Integrating techniques such as incremental learning and anomaly detection \cite{raghuraman2014online}, open-world detectors adapt over time \cite{wang2023detecting}, continuously learning from new data \cite{liu2017lifelong} without forgetting previous knowledge \cite{zhu2024advancing}. This approach is crucial for applications in military surveillance and autonomous navigation, where encountering novel objects is common and can critically impact the decision-making.

Large vision-language models (LVLMs) \cite{zhang2024vision}, which integrate advanced natural language processing with advanced computer vision, are being increasingly utilized in ATR systems \cite{du2022survey}. Models such as CLIP \cite{radford2021learning, hafner2021clip} leverage vast amounts of visual and textual data to enhance object recognition, enabling them to process complex image queries and generate contextually relevant responses. This capability significantly improves detection accuracy and robustness against adversarial attacks or challenging environmental conditions \cite{wu2024safety}, making them highly effective in both military and civilian applications. However, LVLMs face limitations that impact practical deployment. Their detection accuracy often declines in complex scenes or when objects have overlapping features \cite{chen2023applications}, and their performance is sensitive to object size and scale, leading to inaccuracies when targets vary dramatically in size or distance. Furthermore, LVLMs' performance vary based on the prompting method used \cite{nasiriany2024pivot}, making consistent results hard to achieve in critical tasks.

\begin{figure}[tb!]
	\centering
	\includegraphics[width=.9\linewidth]{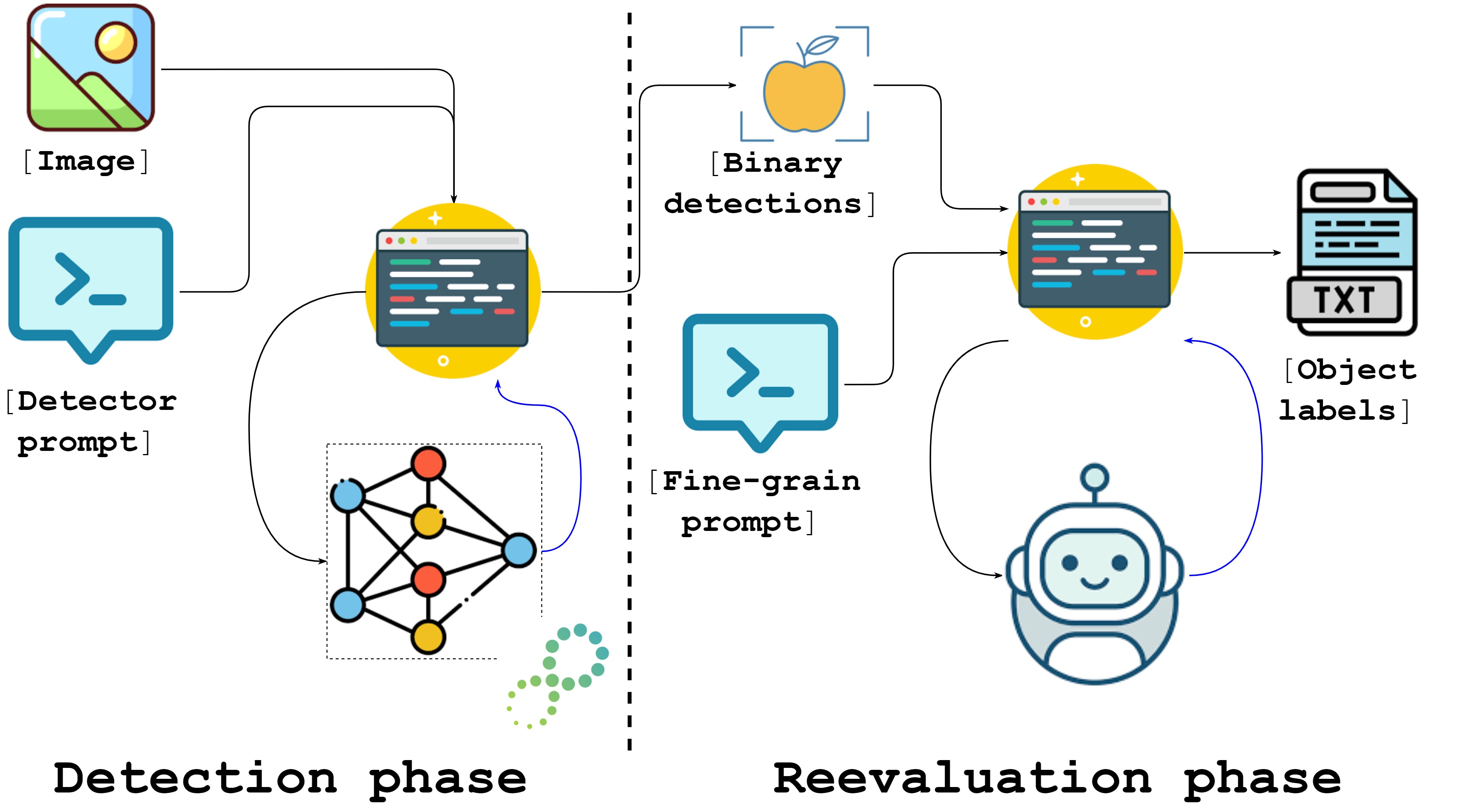}
    \vskip-10pt
    \caption{The proposed for ATR using LVLMs. First, in the `Detection phase,' the image is passed through the object detector for binary detection, where the objects in the scene are detected to produce crops. Then, these crops are sent to the LVLM to recognize the object label in the `Reevaluation phase.'
	}
	\label{fig: pipeline}
\end{figure}

In this work, we focus on leveraging the inherent capabilities of object detectors and LVLMs to perform ATR. LVLMs, with their extensive parameterized memory, can provide more detailed, fine-grained information about a scene or object, despite being less effective at accurately detecting object boundaries in an image. Conversely, current open-world detectors \cite{cheng2024YOLO} excel at localizing objects within a scene, even when the objects belong to novel categories, but they often struggle to correctly classify these objects. Therefore, our approach integrates LVLMs with object detection networks to enhance ATR, especially for novel object classes and domains. We propose a pipeline that generates detection bounding boxes and class labels for objects present in a scene in a zero-shot manner. Furthermore, we study the behavior of LVLMs against factors such as prompting mechanism, image degradation, modality transition, and range effect to understand the limits of the proposed pipeline for ATR.

In summary, this paper makes the following contributions. 1) We introduce a zero-shot pipeline for ATR, leveraging the vast world knowledge embedded within LVLMs. Our approach enables zero-shot object detection and recognition for novel and unseen object classes across diverse environments. 2) We conduct comprehensive experiments, providing insights into the behavior of LVLMs under various prompting strategies to improve zero-shot understanding for ATR applications. 3) We systematically study the impact of critical factors such as image scale and modality on the performance of LVLMs in ATR, providing guidance for optimized ATR deployment in the real-world scenarios.

\section{Related Work}
\noindent{\bf{Open-world ATR}}  is an evolving area of research \cite{bendale2015towards}  that addresses the challenge of detecting and identifying objects in complex and unconstrained environments where the objects may belong to novel categories that the model has not encountered during training as illustrated in Fig. \ref{fig: headline}. Unlike traditional closed-set ATR, which assumes a fixed set of known object classes, open-world ATR must adapt to new and unknown objects in real time. Recent research focuses on leveraging deep learning models, especially those incorporating vision-language models such as CLIP \cite{gupta2022ow}, to improve the detection of novel objects in open-world settings. For instance, \cite{luo2024exploring} highlights the use of vision-language models to generate semantic embeddings for novel object categories, allowing the system to better recognize unseen objects by understanding their relationship to known categories. Moreover, \cite{iqbal2023improved, dhamija2021self} introduce open-world object detectors that focus on localizing and classifying objects from both seen and unseen categories using self-supervised learning approaches. These advancements are pushing the boundaries of ATR by allowing systems to operate in dynamic, real-world environments without the limitations of pre-defined class labels.

\noindent{\bf{Foundation models}} provide an unlimited potential for learning open-world knowledge \cite{narayan2022can}. The effectiveness of the data used in training is crucial to improving the performance of downstream tasks. Segmentation foundation models like SAM \cite{kirillov2023segment} represent a significant leap forward in precise image segmentation, which facilitates zero-shot object recognition. The availability and training on web-scale datasets \cite{melnik2010dremel} have led to the development of increasingly powerful foundation models capable of harnessing vast open-world data. These advancements open new avenues for more intelligent and adaptable systems in various domains.

\section{Proposed Pipeline}
The proposed pipeline for ATR for unseen object classes and novel environmental conditions is illustrated in Fig. \ref{fig: pipeline}. The pipeline is a cascaded two-stage process where a detection module and an LVLM module are combined. Here, in the \textit{Detection} phase, the detection module is used to perform a binary detection which will locate the object crops from the scene. Then, these crops are passed through the LVLM to label during the \textit{Reevaluation} phase.

\subsection{Detection phase}
In the detection module, we intend to crop out objects that are present in the scene of the image. This is because current LVLMs are not able to localize or in other words estimate the bounding box for the objects present in the image, even though LVLMs have good special reasoning capabilities. Furthermore, current state-of-the-art object detectors are still better at producing the bonding box parameters. Hence, in our pipeline, we use the YOLO-world \cite{cheng2024YOLO} object detector as the detection module to perform a binary detection. Here, binary detection refers to simply producing the bounding boxes of the object present in the image without considering the object class. This is because, for novel classes classifier network of the object detector will not produce the correct label. This is why we used YOLO-world as the detector because it allows keyword-prompted object detection. 

To produce the bounding boxes from the detector, we need to provide the keyword or object classes that should be recognized and localized by the YOLO-world pipeline. However, providing the class labels for the detector works mostly in the case of known classes and similar scene domains as the image features are mostly aligned with text embeddings of these known class labels that were available during training. Therefore, the detector will only have high confidence values for the objects of known classes and novel object classes will be removed due to having low confidence or similarity with text embeddings. Besides that, the detector does not know the labels of the unknown or novel classes which precludes the ability to provide novel keywords to prompt the detector. An alternative to this issue is to use an agent that could recognize the objects present in an image scene to provide a list of keywords to prompt the detector. A LVLM is such an entity that contains more world knowledge compared to specialized downstream task networks.

However, we observed in certain cases that the LVLMs fail to recognize certain objects present in the scenes due to the scale of the object compared to the image regardless of whether the object belongs to known or unknown class and novel environment conditions. Nonetheless, the LVLMs were able to provide the labels of the objects that were from the unknown classes that were not present in the original keyword list of the detector. Although we could provide the text prompts for the unknown objects, the detector produced very low confidence scores as these keyword embeddings are not optimized for object recognition. Interestingly, we observed that for unknown or novel objects, even when we provide a similar label to the true class label, or better yet a wrong label, the detector is capable of estimating the bounding boxes but with very low confidence values that are in the range of second or third decimal place. Hence, as a design strategy, we chose to use the keyword `vehicle' to prompt the detector as we are not interested in the classification performance of the detector itself, rather we collect the bounding boxes localized for all the movable objects present in the image. Since we only use a single keyword there are no multiple classes present in the scene and the detector either recognizes or misses the object present in the image thus resulting in a binary detection.

\subsection{Reevaluation phase}
In the reevaluation phase of a detection pipeline, the identified objects undergo labeling, a critical step for ensuring the system’s accuracy and enhancing its performance. To achieve this, LVLMs are increasingly utilized due to their ability to bridge visual and textual data. These models combine the strengths of both computer vision and natural language processing, allowing them to interpret visual content in a semantically rich way. They can understand the context of the detected objects, generate descriptive labels, and even disambiguate objects that may be visually similar but contextually distinct. The primary advantage of using LVLMs lies in their ability to leverage vast amounts of pre-trained data, improving the precision of object identification and labeling. Moreover, these models can handle complex visual scenes by linking images with relevant textual descriptions, making them especially useful for applications where nuanced interpretation of visual data is crucial, such as in autonomous systems. By deploying LVLMs in this reevaluation phase, the labeling process becomes more accurate, context-aware, and scalable. In this work, we study the performance of target recognition under three different three methods: open-set, closed-set, and Chain-of-Thought recognition.

\noindent{\bf{Open-set recognition}}: In open-set recognition \cite{safaei2023open}, the task requires the LVLM to label objects without any prior knowledge of predefined labels. This scenario tests the model's ability to generate meaningful and accurate labels based solely on visual input. Here we give the prompt,

\parbox{0.98\linewidth}{\justify
\texttt{Name the specific vehicle with a single response.}
}

\vspace{1.5mm}
\noindent encouraging it to identify and label the object in the image independently. To assess the effectiveness of the model's recognition performance, we rely on the classification accuracy of the labels it assigns. Upon evaluating the results, we found that the labels generated during the reevaluation phase often differed from the ground truth labels. To address this discrepancy and reconcile the model’s output with the ground truth, we adopted a strategy of selecting the most recurring keyword from the reevaluation labels corresponding to each ground truth class. This method allowed us to bridge the gap between the model’s predictions and the true classifications, offering a more consistent alignment between the two. This approach highlights the potential limitations of open-set recognition but also provides a mechanism to improve accuracy through keyword analysis.

\begin{figure}[!t]
    \centering
    \includegraphics[width=0.38\linewidth]{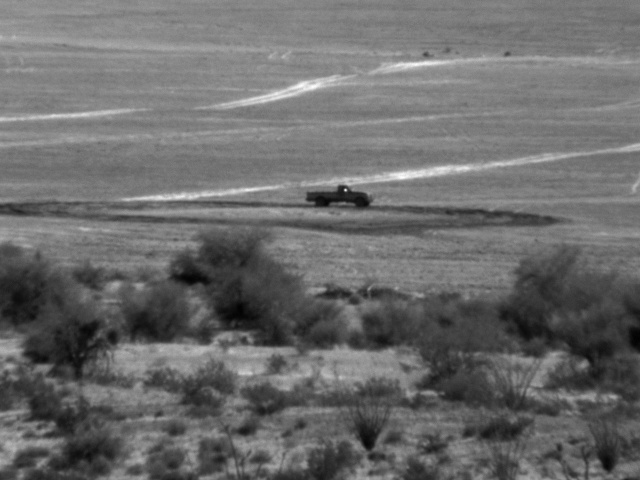}
    \includegraphics[width=0.38\linewidth]{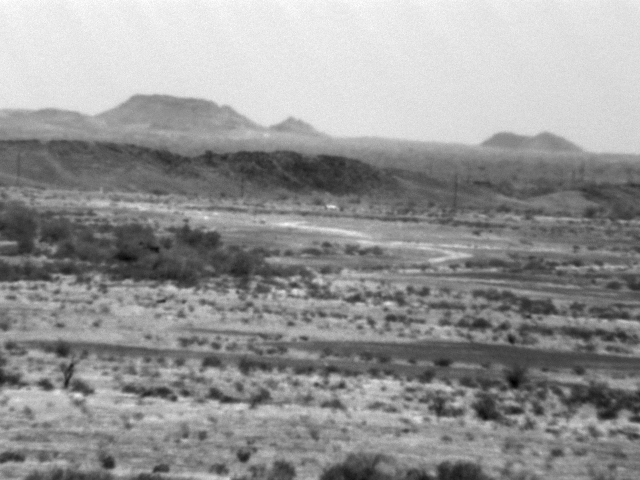}
    
    \vspace{0.15cm}

    \includegraphics[width=0.38\linewidth]{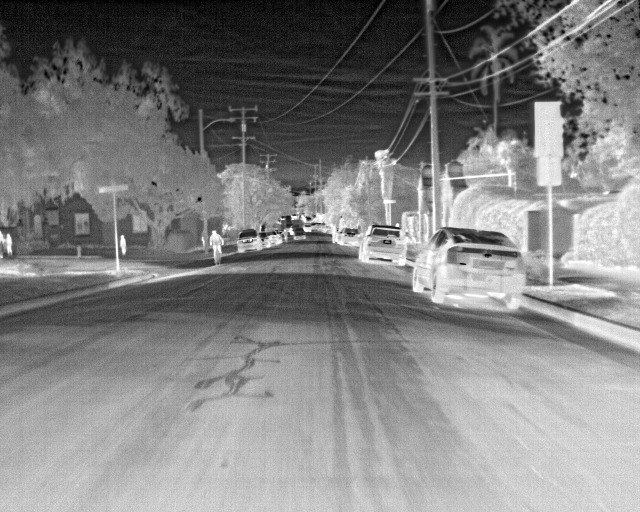}
    \includegraphics[width=0.38\linewidth]{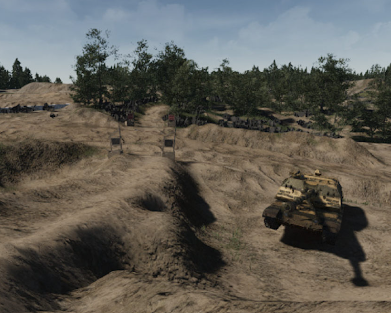}
    
    \vskip-10pt\caption{Sample images from the datasets depicting differences between the conditions tested for automatic target recognition. Left top: near object from the DSIAC dataset with clear visibility. Right top: far object from the DSIAC dataset with difficult visibility. Bottom left: thermal image from ADAS dataset illustrating the deviation from natural images. Bottom right: sample synthetic image from AIS dataset for OOD samples.}
    \label{fig:sample_images}
\end{figure}

\noindent{\bf{Closed-set recognition}}: In closed-set recognition, the LVLM is provided with a set of known labels, which allows the system to operate within predefined constraints. The aim of this approach is to examine how the model's labeling performance is influenced by its awareness of plausible class labels, particularly whether it exhibits bias or changes its responses when given such information. For the closed-set recognition, we use the prompt,

\parbox{0.98\linewidth}{\justify
\texttt{Select a label for the object from the list [known labels, novel]. No long response. Only a single word.}
}

\vspace{1.5mm}
\noindent To handle cases where the detected object does not match any of the known labels, we introduce an additional class labeled as "novel," ensuring that the model has a way to account for unfamiliar objects. In practice, the model is given a prompt such as, `Choose the specific name of the vehicle from the list,' and tasked with selecting the correct label from the provided options. Notably, in this study, the known labels used in the closed-set analysis are drawn from novel object classes that are not included in the classifier network of the detector. This setup allows us to investigate the model’s ability to recognize novel objects within a limited framework and assess how introducing a predefined set of labels impacts its recognition accuracy and decision-making process.

\noindent{\bf{Chain-of-Thought recognition}}: Chain-of-Thought (CoT) recognition is employed to delve into the reasoning process of a vision-language model when selecting labels, allowing us to understand the model’s decision-making pathway. This method is applied in both closed-set and open-set scenarios to explore how the model uses logical steps to arrive at its final label. Furthermore, for CoT recognition we use the following prompt of,For the closed-set recognition, we use the prompt,

\parbox{0.98\linewidth}{\justify
\texttt{Describe the attributes of the vehicle in the image. Build a chain-of-thought to recognize the vehicle. Label the vehicle using the attributes. Give a single word response for label.}
}

\vspace{1.5mm}
\noindent In CoT recognition, the model is first prompted to describe the attributes of the object in the image, such as its shape, color, or other distinguishing features. Based on this description, the model then attempts to recognize and label the object. For open-set labeling, CoT recognition helps measure the reproducibility of the model’s recognition process, ensuring that the model consistently uses the same reasoning pathway to identify objects, even without prior knowledge of known labels. In closed-set labeling, CoT is used to help the model recognize novel objects by drawing on the attributes of the object to select from the predefined set of labels. Performance evaluation for CoT recognition in both open and closed-set cases focuses on the model’s classification accuracy, ensuring that the reasoning process not only makes sense but also leads to accurate label selection. This approach enables a deeper understanding of how the vision-language model processes information and whether its reasoning aligns with human-like cognitive patterns.

\section{Experimental Settings}

\subsection{Datasets}
\noindent{\bf{ADAS dataset}}: The ADAS Dataset was developed to facilitate research in the area of visible and thermal sensor fusion algorithms (commonly referred to as "RGBT") and to support the automotive industry in designing safer and more efficient Advanced Driver Assistance Systems (ADAS) and driverless vehicles. It contains a total of 26,442 fully annotated frames, providing 520,000 bounding box annotations across 15 diverse object categories, including vehicles like cars, trucks, and motorcycles, as well as other objects such as pedestrians, traffic lights, and street signs. The dataset comprises 9,711 thermal and 9,233 RGB images, with a recommended split for training and validation. This dataset is especially valuable for analyzing ATR capabilities within the thermal domain, where traditional RGB sensors fail. 

\noindent{\bf{DSIAC dataset}}: The DSIAC dataset is a specialized collection of monocular images, consisting of 2,595 images designed to support the evaluation of object recognition systems in military contexts. The dataset contains images of vehicles from eight classes captured from varying distances, ranging from 1,000 meters to 5,000 meters, which introduces significant challenges in detecting and classifying objects as the visual clarity diminishes with distance. The class labels in this dataset are specific to military vehicle categories, making it an essential resource for developing and testing recognition algorithms focused on defense applications. The DSIAC dataset serves as an important benchmark for advancing the capabilities of target recognition systems in scenarios where accurate identification at long distances is critical, such as in surveillance, reconnaissance, and autonomous defense operations.

\noindent{\bf{AIS dataset}}: The AIS dataset is a synthetic dataset created using the Applied Intuition Simulator, specifically designed to generate images that simulate desert terrain environments. It contains 200 test images with five general vehicle classes and three military classes, which are intended for zero-shot evaluation of both general and military object categories in novel domains. The use of synthetic images provides flexibility in simulating diverse and challenging environments, such as desert landscapes, where object recognition can be more difficult due to factors like heat distortion, sand, and varying lighting conditions. 
To evaluate the robustness of these models under challenging conditions, weather degradation in the form of simulated rain is applied to the test images. In Fig. \ref{fig:sample_images}, we provide sample images from the datasets.

\subsection{Vision-language models}
In our experiments we use the following LVLMs. For API models: GPT-4o \cite{gpt4o_hello_nodate}, Claude-3.5-Sonnet \cite{anthropic}, and Gemini-1.5-Pro \cite{reid2024gemini}. For open-source models: LLaVA-1.5-7B \cite{liu2023llava}, Phi-3.5-Vision \cite{abdin2024phi}, MiniCPM-Llama3 \cite{yao2024minicpm}, InternVL2-8B \cite{chen2023internvl}, LLaVA-Next \cite{liu2024llavanext}, CogVLM \cite{wang2023cogvlm}, OpenFlamingo-v2 \cite{awadalla2023openflamingo}, InstructBLIP \cite{instructblip}, BLIP2 \cite{li2023blip} and as the baseline model CLIP \cite{radford2021learning}.

\section{Results}

The performance of the proposed pipeline for different LVLMs for the datasets are tabulated in 
Table \ref{table: adas} for the ADAS dataset, Table \ref{table: ais} for the AIS dataset, Table \ref{table: dsiac} for the DSIAC dataset, and Table \ref{table:weather} for weather degradation.
Generally, the API models perform better by a significant margin as these models are bigger models compared to the other open-source models. However, the open-source LVLMs perform comparatively well compared to other smaller models (CLIP) that are generally used in open-world detectors. 

\begin{table}[!t]
	\centering
\caption{Model performance comparison on ADAS dataset.}
	\vskip-10pt
	\begin{tabular}{@{}l c c c c@{}}
\toprule
\multirow{2}{*}{Model} & \multirow{2}{*}{Open-set} & \multirow{2}{*}{Closed-set} & \multicolumn{2}{c}{Chain-of-Thought} \\ 
                            &   &   & Open-set  & Closed-set \\ \hline
GPT-4o& 58.23 & 60.11 & 59.13 & 63.06 \\
Claude-3.5-Sonnet           & 57.85 & 58.24 & 59.02 & 64.94 \\
Gemini-1.5-Pro              & 54.61 & 56.83 & 61.11 & 62.86 \\[1.5ex]
LLaVA-1.5-7B                       & 35.63 & 37.87 & 42.82 & 40.39 \\
Phi-3.5-Vision              & 39.10  & 40.31  & 41.45  & 46.48  \\
MiniCPM-Llama3              & 33.90  & 34.41  & 38.34  & 39.44  \\
InternVL2-8B                 & 29.50 & 29.60 & 30.19 & 38.29 \\
LLaVA-Next                  & 43.30  & 44.13  & 47.30  & 51.26  \\
CogVLM                      & 49.34 & 51.38 & 55.51 & 58.08 \\
OpenFlamingo-v2             & 22.83 & 24.10 & 24.28 & 25.85 \\
InstructBLIP                & 17.13 & 18.72 & 20.89 & 23.23 \\
BLIP2                       & 10.92 & 11.59 & 15.41 & 15.83 \\
CLIP                        & 5.43  & 6.37  & 10.65 & 10.33 \\
\bottomrule
\end{tabular}
	\label{table: adas}
\end{table}

\begin{table}[!t]
	\centering
\caption{Model performance comparison on AIS dataset.}
	\vskip-10pt
	\begin{tabular}{@{}l c c c c@{}}
\toprule
\multirow{2}{*}{Model} & \multirow{2}{*}{Open-set} & \multirow{2}{*}{Closed-set} & \multicolumn{2}{c}{Chain-of-Thought} \\ 
                            &   &   & Open-set  & Closed-set \\ \hline
GPT-4o                      & 64.20 & 64.27 & 64.42 & 69.03 \\
Claude-3.5-Sonnet           & 63.84 & 66.30 & 68.96 & 73.29 \\
Gemini-1.5-Pro              & 60.62 & 63.10 & 66.67 & 66.24 \\[1.5ex]
LLaVA-1.5-7B                       & 41.63 & 42.06 & 45.76 & 45.90 \\
Phi-3.5-Vision              & 45.10  & 45.65  & 50.38  & 49.76  \\
MiniCPM-Llama3              & 39.90  & 41.77  & 45.14  & 44.09  \\
InternVL2-8B                & 35.50 & 36.05 & 40.69 & 40.20 \\
LLaVA-Next                  & 49.30  & 52.05  & 54.75  & 55.42  \\
CogVLM                      & 55.34 & 57.69 & 60.72 & 64.34 \\
OpenFlamingo-v2             & 28.82 & 30.89 & 35.83 & 35.81 \\
InstructBLIP                & 23.13 & 24.08 & 25.40 & 33.07 \\
BLIP2                       & 16.94 & 17.17 & 19.83 & 25.70 \\
CLIP                        & 11.48 & 11.62 & 13.64 & 14.62 \\
\bottomrule
\end{tabular}
	\label{table: ais}
\end{table}

\begin{table*}[!t]
	\centering
\caption{Model performance comparison on DSIAC dataset.}
	\vskip-10pt
	\begin{tabular}{@{}l c c c c c c c c c c@{}}
\toprule
Range (m) & \multicolumn{4}{c}{1000} & \multicolumn{4}{c}{2000} & \multicolumn{2}{c}{3000-5000} \\\cmidrule(lr){2-5}\cmidrule(lr){6-9}\cmidrule{10-11}
\multirow{2}{*}{Model} & \multirow{2}{*}{Open-set} & \multirow{2}{*}{Closed-set} & \multicolumn{2}{c}{Chain-of-Thought} & \multirow{2}{*}{Open-set} & \multirow{2}{*}{Closed-set} & \multicolumn{2}{c}{Chain-of-Thought} & \multicolumn{2}{c}{Chain-of-Thought}\\
                            &        &        & Open-set  & Closed-set &   &   & Open-set  & Closed-set & Open-set  & Closed-set\\ \hline
GPT-4o                        & 67.23  & 69.32  & 68.48  & 73.61  & 66.94  & 67.03  & 69.17  & 73.45  & 34.22 & 36.39   \\
Claude-3.5-Sonnet                      & 66.82  & 67.65  & 68.32  & 73.71  & 63.55  & 64.49  & 66.06  & 70.37  & 31.34 & 32.78   \\
Gemini-1.5-Pro                    & 63.62  & 64.93  & 65.18  & 72.52  & 62.29  & 65.00  & 63.64  & 66.14  & 28.17 & 30.70   \\[1.5ex]
LLaVA-1.5-7B                        & 44.63  & 46.71  & 48.75  & 49.04  & 42.81  & 44.35  & 46.20  & 46.71  & 15.13 & 17.14   \\
Phi-3.5-Vision                       & 54.86  & 56.79  & 56.24  & 51.50  & 52.90  & 53.65  & 59.40  & 41.76  & 16.79 & 18.04  \\
MiniCPM-Llama3                     & 48.44  & 48.54  & 53.93  & 46.09  & 47.32  & 47.35  & 52.99  & 41.35  & 14.17 & 15.48  \\     
InternVL2-8B                     & 44.45  & 47.26  & 50.11  & 40.36  & 42.62  & 40.38  & 43.93  & 37.75  & 16.45 & 17.74  \\
LLaVA-Next                       & 57.28  & 60.82  & 61.48  & 53.78  & 56.26  & 57.52  & 59.99  & 48.82  & 15.54 & 17.19  \\  
CogVLM                             & 58.34  & 60.84  & 62.52  & 68.05  & 54.92  & 55.65  & 57.74  & 57.95  & 18.77 & 20.68   \\
OpenFlamingo-v2          & 31.83  & 33.44  & 35.24  & 35.31  & 29.58  & 30.92  & 30.44  & 33.35  & 11.23 & 14.09   \\
InstructBLIP                         & 26.13  & 26.80  & 31.67  & 32.34  & 25.76  & 27.00  & 25.98  & 33.46  & 9.68  & 11.68   \\
BLIP2                                  & 19.94  & 21.13  & 21.49  & 28.55  & 18.50  & 19.06  & 18.60  & 22.21  & 1.61  & 2.27     \\
CLIP                          & 14.43  & 17.12  & 20.01  & 18.40  & 10.26  & 10.86  & 10.30  & 14.21  & 1.06  & 1.43        \\
\bottomrule
\end{tabular}
	\label{table: dsiac}
\end{table*}

\begin{figure}[!t]
    \centering
    \includegraphics[width=0.45\linewidth]{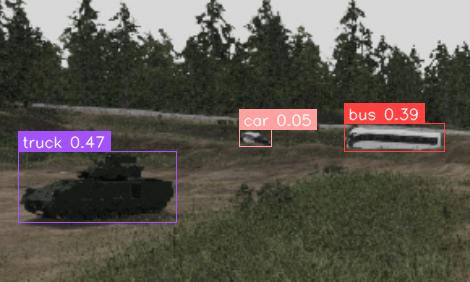}
    \includegraphics[width=0.45\linewidth]{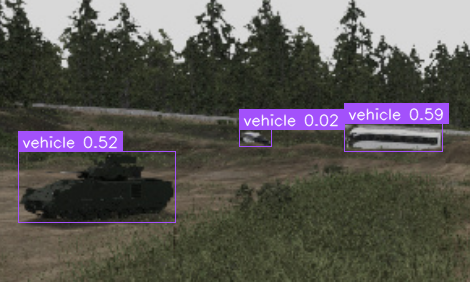}
    
    \vspace{0.15cm}    
    
    \includegraphics[width=0.45\linewidth]{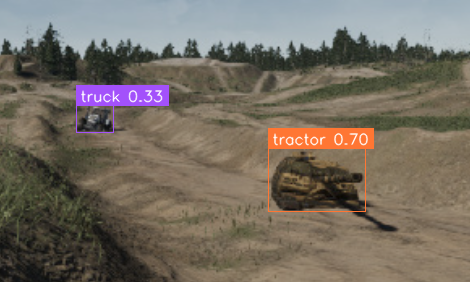}
    \includegraphics[width=0.45\linewidth]{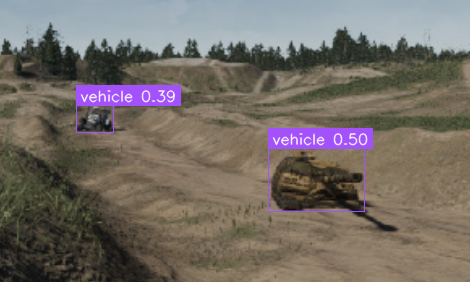}
    
\vskip-10pt    \caption{Misrecognition by open-world detectors for novel object categories (first column) and the localization performance of binary detection (second column) compared to using a keyword vocabulary.}
    \label{fig:binary_detection}
\end{figure}

\noindent {\bf{Effect of binary detection.}}
In Fig. \ref{fig:binary_detection}, we illustrate the detection performance of binary detection alongside keyword detection. For keyword detection, we provided the keywords used by the YOLO-world pipeline and supplemented them with additional keywords extracted from a LVLM based on the image scene. As can be seen from the first column in Fig. \ref{fig:binary_detection}, for novel class objects such as the tank and tractor, the YOLO-world detector was unable to recognize or classify them with correct labels. This highlights the limitations of current open-world detectors in automatic recognition. Furthermore, performing binary detection produced the same level of localization performance as keyword detection, which validates the decision to remove the object vocabulary for localization. Additionally, we observed that, for the same image, using different keywords altered the confidence scores of the classifications, unlike binary detection. This variation makes it challenging to set a confidence threshold to filter out localization results with very low confidence scores.

\begin{figure}[!t]
    \centering
    \includegraphics[width=0.45\linewidth]{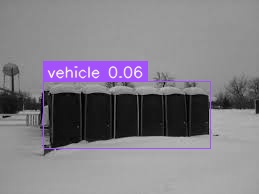}
    \includegraphics[width=0.45\linewidth]{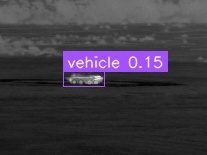}

    \vspace{0.15cm}    
    
    \includegraphics[width=0.45\linewidth]{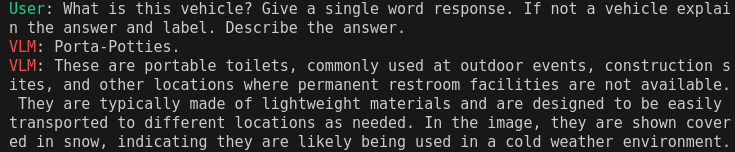}  
    \includegraphics[width=0.45\linewidth, height=0.87cm]{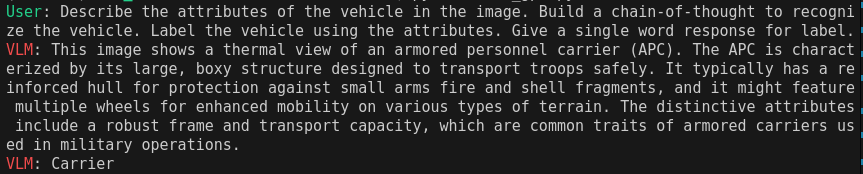}
    
\vskip-10pt    \caption{The pipeline can be used to remove false positives (left image) produced by the detector. The Chain-of-thought recognition on the thermal image illustrates the attributes used to label the object.}
    \label{fig:false_positives}
\end{figure}

\noindent {\bf{Removing false positives.}}
In most detection pipelines, false positives are frequently captured, which significantly reduces detection performance. This presents a major limitation, as there is no way to directly remove the false positives from the trained model. Such inaccuracies are particularly dangerous when safety is a primary concern, as wrong decision-making based on these false positives could lead to negative consequences. However, LVLMs can be used to verify and filter the captured objects. As shown in Fig. \ref{fig:false_positives} (left image), the pipeline can effectively remove false positives produced by the detector. This practice can be extended to general ATR systems to help eliminate false detections.

\noindent {\bf{Chain-of-Thought recognition.}}
In Fig. \ref{fig:false_positives}, we provide an example of CoT recognition in the thermal domain. We observed an increase in recognition performance using the CoT method for both open and closed-set recognition. Specifically, for the example in Fig. \ref{fig:false_positives}, the pipeline initially recognized the object as a tank under open set recognition, despite it being a thermal image of an armored personnel carrier. With the CoT approach, the pipeline was able to correctly recognize the military vehicle as a carrier by utilizing descriptions of the object as secondary input. This capability is unique to LVLMs, as traditional detectors are not able to identify the attributes of the vehicle when classifying.

\begin{figure}[!b]
    \centering
    \parbox[b]{0.32\linewidth}{
        \centering
        \includegraphics[width=\linewidth]{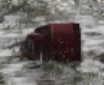}
        \scriptsize
        \raggedright
        GT: Tank \\[-1pt]
        GPT-4o: Tank \\[-1pt]
        LLaVA-Next: Tank
    }
    \hfill
    \parbox[b]{0.32\linewidth}{
        \centering
        \includegraphics[width=\linewidth]{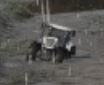}
        \scriptsize
        \raggedright
        GT: Tractor \\[-1pt]
        GPT-4o: Tractor \\[-1pt]
        LLaVA-Next: ATV
    }
    \hfill
    \parbox[b]{0.32\linewidth}{
        \centering
        \includegraphics[width=\linewidth]{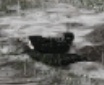}
        \scriptsize
        \raggedright
        GT: Tank \\[-1pt]
        GPT-4o: Boat \\[-1pt]
        LLaVA-Next: None
    }
    
\vskip-7pt    \caption{Qualitative examples for ATR under weather degradation. The smaller models fail to locate heavily distorted scenes, and the larger models fail to properly recognize the object.}
    \label{fig:weather_degrade}
\end{figure}
 
\noindent {\bf{Effect of image degradation.}}
In Fig. \ref{fig:weather_degrade} and Table \ref{table:weather}, we present the performance of different vision-language models under adverse weather conditions. The recognition ability generally drops, as expected, since the quality of the scene is hindered. This occurs because the attributes of the objects present in the image under these conditions can differ from those in a clear scene. In Fig. \ref{fig:weather_degrade}, the labels given by the best-performing model for the dataset were truck, tractor, and boat, whereas the ground truth was truck, tractor, and tank, from left to right. The smaller models misidentified the tractor as an ATV and failed to recognize the tank as any vehicle. This aligns with human recognition capabilities, as the tractor shares a similar structure with an ATV, and the scene with the tank was severely degraded. Also, when the CoT method was applied, the decision-making involved considering the background of the scene. For example, the label `boat' was assigned to the tank during reevaluation, with the explanation: \textit{`The curved hull and the surrounding water indicate that this vehicle is designed for water navigation. The dark silhouette contrasts with the lighter water around it, emphasizing the shape typical of a boat.'}

\begin{table}[!t]
	\centering
\caption{Model performance under weather degradation.}
	\vskip-10pt
	\begin{tabular}{@{}l c c c c@{}}
\toprule
\multirow{2}{*}{Model} & \multirow{2}{*}{Open-set} & \multirow{2}{*}{Closed-set} & \multicolumn{2}{c}{Chain-of-Thought} \\ 
                            &   &   & Open-set  & Closed-set \\ \hline
GPT-4o                      & 61.24 & 63.52 & 61.92 & 70.92 \\ 
Claude-3.5-Sonnet           & 60.88 & 60.83 & 65.43 & 66.14 \\ 
Gemini-1.5-Pro              & 57.63 & 60.36 & 64.97 & 64.71 \\ [1.5ex]
LLaVA-1.5-7B                     & 38.67 & 39.43 & 43.73 & 42.71 \\ 
Phi-3.5-Vision              & 42.10  & 44.11  & 46.13  & 49.70  \\
MiniCPM-Llama3              & 36.90  & 39.07  & 39.33  & 40.81  \\
InternVL2-8B                  & 32.50 & 34.46 & 35.98 & 38.25 \\
LLaVA-Next                  & 46.30  & 47.83  & 50.00  & 50.92  \\
CogVLM                       & 52.31 & 52.73 & 53.42 & 58.03 \\ 
OpenFlamingo-v2             & 25.82 & 28.33 & 29.88 & 33.30  \\ 
InstructBLIP                & 20.12 & 20.99 & 23.4  & 28.19 \\ 
BLIP2                       & 13.98 & 14.36 & 18.09 & 19.72 \\ 
CLIP                        & 8.45  & 9.54  & 13.29 & 15.59 \\ 
\bottomrule
\end{tabular}
	\label{table:weather}
\end{table}

\noindent {\bf{Future work.}}
While these large models contain extensive world knowledge, they are challenging to apply in real-time applications due to time constraints. However, they can serve as excellent teacher models or agents to train smaller, specialized models, where the large models can disseminate knowledge about novel objects encountered by the specialized models. Furthermore, the recognition performance of LVLMs significantly surpassed that of theYOLO-world detector for both RGB and grayscale images. However, for thermal images, the performance improvement was not as substantial as with other modalities. Therefore, these large models can be fine-tuned or adapted to other domains, such as thermal imaging, to enhance open-world target recognition across domains or even towards unified models.

\section{Conclusion}
In conclusion, our work demonstrated the use of LVLMs for zero-shot ATR in novel environments and object categories. We showed that by combining the detection capabilities of existing object detectors with the world knowledge of LVLMs, we can overcome the performance drop in open-world detectors for extreme novel object classes or environments, while also addressing the poor localization capabilities of LVLMs. This work highlights how these foundation models can be employed to develop more reliable systems where safety and accuracy are of paramount importance. Additionally, we presented the performance of the proposed pipeline with different LVLMs for comparison across various modalities and conditions. Furthermore, we emphasized key advantages, such as false positive removal, binary detection, and Chain-of-Thought recognition, made possible by our pipeline through the use of LVLMs. By providing future directions, we hope that our work will shed light on new approaches for ATR in the era of LVLMs, thereby facilitating advancements in the right direction.

\bibliographystyle{IEEEtran}
\bibliography{main}

\end{document}